\documentclass[letterpaper, 10 pt, conference]{ieeeconf}

\IEEEoverridecommandlockouts                    

\usepackage{amsmath} 
\usepackage{amssymb}  
\usepackage{amsfonts}  
\usepackage{graphicx} 
\usepackage[font=small,labelfont=bf,tableposition=top]{caption}
\usepackage{booktabs}
\usepackage{multicol}
\usepackage{subfigure}
\usepackage{cuted}
\usepackage{microtype}
\usepackage{fancyhdr}
\usepackage[colorlinks=true,
        linkcolor=black,   
        citecolor=black,   
        filecolor=black,
        urlcolor=black,
        bookmarks=true,
        bookmarksopen=true,
        bookmarksopenlevel=3,
        plainpages=false,
        pdfpagelabels=true]{hyperref}
\usepackage[all]{hypcap}
\usepackage{microtype}
\usepackage{cite}
\usepackage{rotating}

\graphicspath{ {images/} } 

\usepackage{xcolor}

\newcommand{\etal}{\textit{et al.}}

\newcommand{\vect}[1]{\mathbf{ #1}}

\newcommand{\argmax}{\operatornamewithlimits{argmax}}

\newcommand{\vs}{\vect{s}}

\newcommand{\cC}{\mathcal{C}}

\title{\LARGE \bf
Meaningful Maps With Object-Oriented Semantic Mapping
}

\author{
Niko S\"underhauf$^{1}$,             
Trung T. Pham$^{2}$,                 
Yasir Latif$^{2}$,                   
Michael Milford$^{1}$,               
Ian Reid$^{2}$                       
\thanks{This research was supported by the Australian Research Council Centre of Excellence for Robotic Vision, project number CE140100016. Michael Milford is supported by an Australian Research Council Future Fellowship FT140101229.}
\thanks{The authors are with the Australian Centre for Robotic Vision.}%
\thanks{$^{1}$ Niko and Michael are with Queensland University of Technology (QUT), Brisbane, QLD 4001 Australia.}%
\thanks{$^{2}$ Trung, Yasir, and Ian are with the University of Adelaide, SA 5005 Australia.}%
\thanks{Contact: {\tt\small niko.suenderhauf@roboticvision.org}}
}

\begin{document}

\maketitle

\thispagestyle{empty}

\begin{strip}

  \vspace{-2.5cm}
  \centering
  \includegraphics[width=0.24\linewidth]{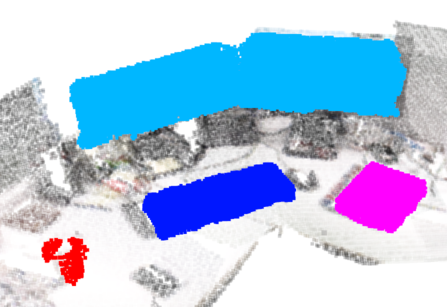}
  \includegraphics[width=0.24\linewidth]{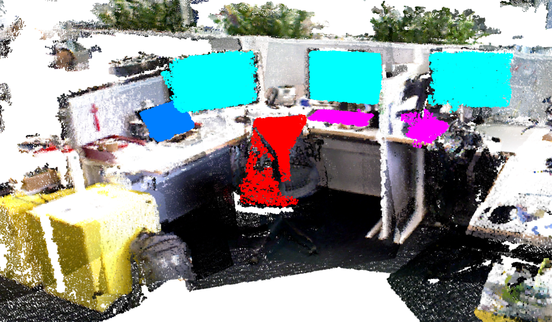}
  \includegraphics[width=0.24\linewidth]{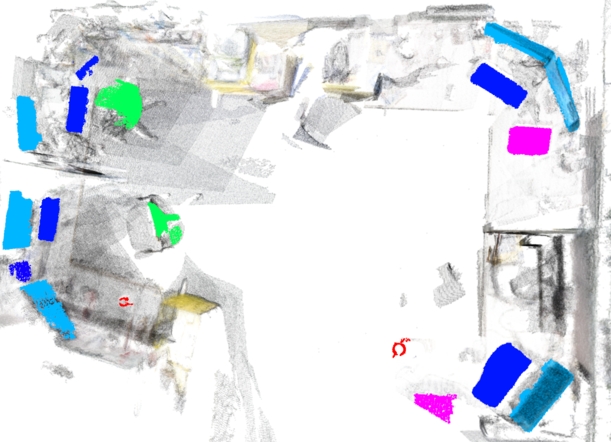}
  \includegraphics[width=0.24\linewidth]{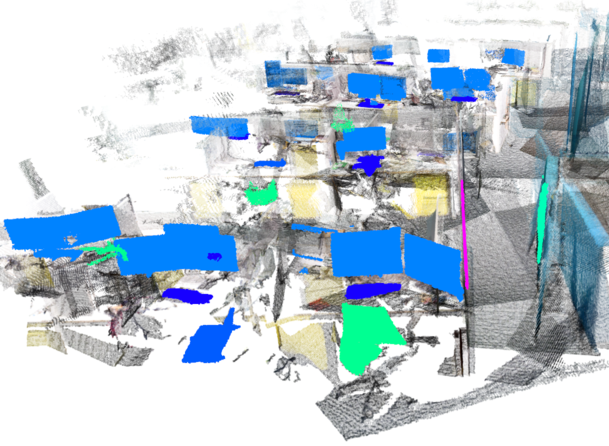}
  \captionof{figure}{We demonstrate object-oriented semantic mapping using RGB-D data that scales from small desktop environments (left) to offices (middle) and whole labs (right). The pictures show 3D map structures with objects colored according to their semantic class. We do not merely project semantic labels for individual 3D \emph{points}, but rather maintain \emph{objects} as the central entity of the map, freeing it from the requirement for \textit{a-priori} 3D object models in \cite{Salas13}. To achieve this, our system creates and extends 3D object models while continuously mapping the environment. Object detection and classification is performed using a Convolutional Network, while an unsupervised 3D segmentation algorithm assigns a segment of 3D points to every object detection. These segmented object detections are then either fused with existing objects, or added as a new object to the map. ORB-SLAM2 provides a global SLAM solution that enables us to reconstruct a 3D model of the environment that contains both non-object structure and objects of various types.}
 \label{fig:example_maps}
\end{strip}
\pagestyle{empty}

\begin{abstract}

For intelligent robots to interact in meaningful ways with their environment, they must understand both the geometric and semantic properties of the scene surrounding them. The majority of research to date has addressed these mapping challenges separately, focusing on either geometric or semantic mapping. In this paper we address the problem of building environmental maps that include both semantically meaningful, object-level entities and point- or mesh-based geometrical representations. We simultaneously build geometric point cloud models of previously unseen instances of known object classes and create a map that contains these object models as central entities. Our system leverages sparse, feature-based RGB-D SLAM, image-based deep-learning object detection and 3D unsupervised segmentation.

\end{abstract}

\section{Introduction}

For mobile robots to interact meaningfully with their environment, they must have access to a world model that conveys both geometric and semantic information. However, most recent research in robotic mapping and SLAM has  concentrated on either accurately modelling only the \emph{geometry} of the world, or focused mapping of a few semantic classes but neglected modelling separate object \emph{instances} or required \textit{a-priori} known 3D object models.

In this paper we present an approach for creating more \emph{meaningful maps} without some of the requirements and limitations of previous approaches: maps that not only express \emph{where} something is in the world, but also \emph{what} it is. 
Our approach is timely in that geometry-focused robotic SLAM and deep-learning-based object detection techniques are now mature enough to be incorporated into an \emph{object-oriented} semantic mapping system that can create richly annotated maps with many dozens or even hundreds of object classes, while maintaining geometric accuracy.

Our semantic mapping is \emph{object-oriented} since individual object \emph{instances} are the key entities in our map. As illustrated in Fig. \ref{fig:example_maps}, the generated map of an environment is enriched with semantic information in the form of separate object entities. These objects carry both geometric and semantic information in the form of class labels and confidences. An important distinction to earlier work (e.g. \cite{Stueckler15}) is that our map does \emph{not} merely maintain labeled \emph{independent} 3D points by projecting semantic information from images into the 3D structure. Rather, the objects are separate entities completely independent from the non-object parts of the map. This enables more advanced scene understanding, e.g. a robot can reason that all 3D points belonging to one object in the map will move together upon manipulation. 

This object-centric approach is supported by an instance-level semantic segmentation technique that combines bounding box-based object detection in the image space with unsupervised 3D segmentation. In contrast to our instance-level approach, semantic segmentation approaches such as \cite{Tateno16, Long15} often have no notion of object instances, and are therefore less usable in a robotic mapping setup where individual objects need to be modeled and maintained in the map over time.

Our approach creates and extends 3D object models as it maps the environment without requiring \textit{a-priori} known 3D models as in \cite{Salas13}. This is a practically significant improvement over previous techniques, since individual instances of a semantic class like \texttt{chair} can vary significantly even within a single environment, and requiring precise 3D models for all of these variations is severely limiting in terms of practical robotic implementations.

In the rest of this paper, we first discuss related work before describing our proposed system in detail. A quantitative and qualitative evaluation demonstrating the semantic mapping system in action is provided in Section \ref{sec:results}.
Finally we discuss the results and insights obtained, followed by a discussion for future research directions in this combined geometric-semantic mapping area.

\section{Related Work}
The geometric aspect of the SLAM problem is well understood and has reached a level of maturity where city level maps can be built precisely and in real time \cite{Mur-Artal15}. However, the outcome of such maps is geometric entities (points, planes, surfaces etc.), which, while useful for the task of mapping and localization, do not inform an active agent of the identity or the list of possible actions that can be carried out on the entities present in the environment. In order to interact with the environment a semantic representation is needed. The granularity of the semantic labels depends on the task at hand. 
A robot that needs to reason about moving from point A to B needs access to place identities (room, corridor, kitchen etc), 
while a robot that manipulates objects needs information about object identities and affordances
(What can be done with the object? How to grab it? How is it supported in space?).

\subsection{Semantic Mapping}
Semantic mapping is the process of attaching semantic meaning (object categories, identities, actions, etc.) to the entities being mapped. 
It uses SLAM as a tool to reason about the motion and position of a sensor in the environment, 
while semantic information may be obtained from a different source. 

One of the first approaches towards semantic mapping involved map reconstruction followed by segmentation of the reconstructed map into semantic concepts \cite{Herbst11, Pillai15, pham2015}. Pham \etal~\cite{pham2015} first reconstruct a dense 3D model from RGB-D images using KinectFusion \cite{Newcombe11}, then assign every 3D point a semantic label using a hierarchical Conditional Random Fields (CRF) model. Pillai and Leonard~\cite{Pillai15} used a monocular SLAM system to boost the performance of object recognition in videos. They showed that the  performance of the object recognition task improved when supported temporally by Monocular SLAM. Herbst \etal \cite{Herbst11} proposed an unsupervised algorithm for object discovery by comparing the 3D models reconstructed using SLAM from multiple visits to the same scene. Unlike these works, our method tracks the camera, detects and reconstructs the object models on the fly.

Other semantic mapping methods involve online scene reconstruction and segmentation \cite{Mozos07,Pronobis12,Hermans14, Pronobis09,Cadena15,Vineet15,Kundu14}. The work of Mozos \etal~\cite{Mozos07} segments maps built with range sensors into functional spaces (rooms, corridors, doorways) using a hidden Markov model. They show that the semantic information thus obtained can then be used to convert the geometric map into a topological map. Pronobis \etal~\cite{Pronobis12,Pronobis09} proposed an online system to build a semantic map of the environment using laser as well as cameras. Cadena \etal~\cite{Cadena15} use the motion estimation combined with an opportunistic distributed system of different object detectors to outperform the individual systems. Vineet \etal~\cite{Vineet15} proposed an online dense reconstruction method that solves the semantic labelling problem as a densely connected CRF. Kundu \etal~\cite{Kundu14} derive a CRF model over 3D space, that jointly infers the semantic category and occupancy for each voxel. Our work, in contrast, models objects as separate entities in space which already informs an active agent of its inherent properties (such as rigid motion upon manipulation etc.), which is not possible with point-wise labelled maps. While one could generate object instances from point-wise labelling maps using a post-processing 3D segmentation algorithm, such an approach is inefficient for online systems.

Semantic information can also be added to the map by object-template matching. Civera \etal~\cite{Civera12} match the map points created using a monocular SLAM system against a known database of objects, which upon recognition using a feature based methods, can be inserted into the map. This creates more complete maps and allows for scale resolution. Similarly, Castle \etal~\cite{Castle07} use planar known objects in a monocular SLAM framework. Salas-Moreno \etal \cite{Salas13} create object based maps, which uses RGB-D information to recognize and insert models of known objects. Our work differs from these methods as there is no prior database of objects. Our method creates object models on the fly based on the output of an object detector.

\subsection{Object Detection and Semantic Segmentation}
\label{sec:related_work_detection}

Geometry alone is ambiguous for the task of object detection and with the current advancement in the field of machine learning, rich priors can be learnt from data itself. Specifically, we are interested in the task of object detection that can be utilized to isolate object instances in the map. One of the methods for object detection is the
\emph{proposal-based object detection} which generates a number $n$ of object \emph{proposals}, typically in the form of bounding boxes. Each of those proposals is classified, resulting in $n$ independent probability distributions over all class labels. This technique was pioneered by R-CNN~\cite{Girshick14} and recently developed further by approaches like Fast~R-CNN~\cite{Girshick15}, Faster~R-CNN~\cite{Ren15}, YOLO~\cite{Redmon15}, or the Single Shot MultiBox Detector~(SSD)~\cite{Liu15}.

As opposed to object detection,
semantic segmentation \cite{Long15} generates dense, pixel-wise classification. 
The disadvantage of such methods for semantic mapping is that they often lack the notion of independent object \emph{instances}. 
Pixel labels for overlapping objects, therefore, do not allow identification of individual objects present in the scene, which can lead to data-association ambiguities in a SLAM framework. Recent work towards instance-level semantic segmentation using rgb images include \cite{Zagoruyko16, Dai15, Zhang16, Uhrig16}.  While there has been significant progress, at present these methods lack the accuracy and the speed to be usable in our online framework.

\section{Object-Oriented Semantic Mapping}

\begin{figure}[t]
  \centering
  \includegraphics[width=0.45\linewidth]{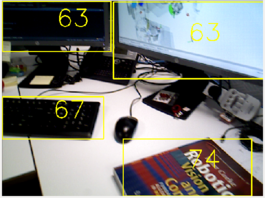}
  \includegraphics[width=0.45\linewidth]{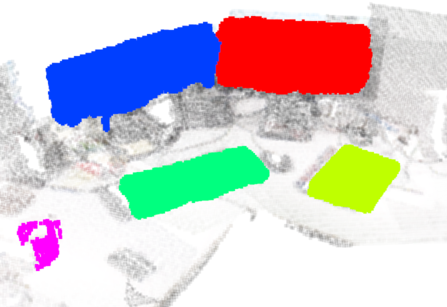}
  \includegraphics[width=0.45\linewidth]{objects_labels.png}
  \includegraphics[width=0.45\linewidth]{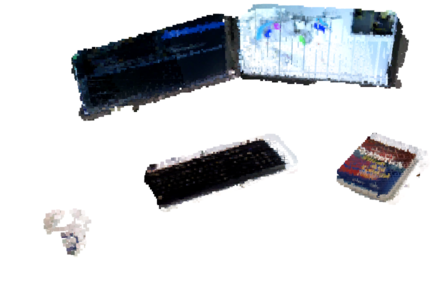}
  \caption{Illustration of key steps in our proposed approach: (top-left) SSD~\cite{Liu15} generates object proposals consisting of bounding boxes, class labels, and confidence scores. (top-right) Our unsupervised 3D segmentation algorithm creates a 3D point cloud segment for each of these objects detected in the current RGB-D frame. (bottom row) We obtain a map that contains semantically meaningful entities: objects that carry a semantic label, confidence, as well as geometric information. The semantic label is color coded in the bottom left image. light blue: monitor, pink: book, red: cup, dark blue: keyboard.}
  \label{fig:system_overview2}
\end{figure}

\begin{figure*}[t]
  \centering
  \includegraphics[width=\linewidth]{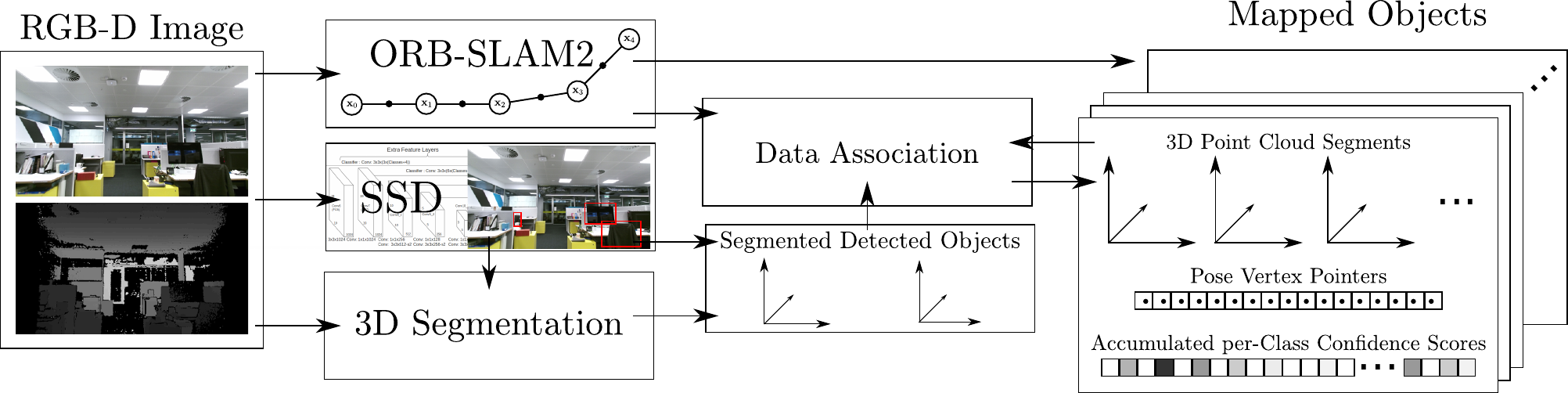}
  \caption{Overview of our semantic mapping system. While ORB-SLAM2 performs camera localisation and mapping on every RGB-D frame, SSD~\cite{Liu15} detects objects in every RGB \emph{keframe}. Our own adapted 3D unsupervised segmentation approach assigns a 3D point cloud segment to every detection. Data association based on an ICP-like matching score decides to either create a new object in the map or associate the detection with an existing one. In the latter case, the 3D model of the map object is extended  with the newly detected 3D structure. Every object stores 3D point cloud segments, pointers into the pose graph of ORB-SLAM and per-class confidence scores that are updated on the fly whenever new observations are available.
}
  \label{fig:system_overview}
\end{figure*}

In this section, we outline the main components that constitute our semantic mapping system.
Fig.~\ref{fig:system_overview2} visualizes key stages of the proposed approach while Fig.~\ref{fig:system_overview} illustrates the flow of  information between the different components.

\subsection{SLAM}

Our semantic mapping system requires knowledge of the current camera pose when observing a scene. 
This information can come from a SLAM system that is running in the background (SLAM as a service) or from localization against a previously built map. 
In this work, we use an out-of-the-box RGB-D version of ORB-SLAM2 \cite{Mur-Artal15} for tracking and mapping, which uses both RGB and depth information for sparse tracking. The point clouds generated from the depth channel are projected to the corresponding sensor location to get a map of the environment.

We are cautious of terming our method ``semantic SLAM'' as that requires information flow in both directions: SLAM helping semantics and semantics helping SLAM. 
In the proposed method, the information flows in only one direction, that is, SLAM helping semantics: 
the output of SLAM (camera poses, map) is used to achieve a coherent semantic labelling of the environment. 
Therefore our work belongs to the category of ``semantic mapping''. 
However, the system can be extended to incorporate bidirectional information flow but currently that is left as future work.

\subsection{Object Detection for Semantic Mapping}
Individual objects are crucial entities in a semantic map. Therefore, a method to localize and recognize different object instances in an image is needed. Although there is considerable progress on instance level semantic segmentation \cite{Zagoruyko16, Dai15, Zhang16, Pinheiro15, Uhrig16}, these works are not sufficiently fast for our semantic mapping framework. For example, DeepMask~\cite{Pinheiro15} takes about 1.6s per image. On the contrary, deep-learning proposal-based object detection approaches have shown excellent results and real-time performance \cite{Girshick14, Girshick15, Ren15, Redmon15, Liu15}.

We use the Single-shot Multi-box Detector (SSD) approach~\cite{Liu15} to generate a fixed number of object proposals in the form of bounding boxes for every keyframe. SSD provides a class label and a confidence score $0\le \mathfrak{s} \le 1$ for every proposal. It has demonstrated highly competitive results on the established computer vision benchmarks MS COCO~\cite{Lin14}, PASCAL VOC, and ILSVRC~\cite{Russakovsky15}.

We use the network trained on the COCO dataset for our work, which can recognize 80 classes. A forward pass through the $500\times 500$ variant of the network, i.e. acquiring proposals and classifications, takes 86 ms on a TitanX GPU.

\subsection{3D Segmentation}

Image-based object detection methods hardly return bounding boxes fitted well to objects (see Fig. \ref{fig:system_overview2} top-left). However, it is necessary to have precise object boundaries for better object models reconstruction. To this end, we leverage depth information to generate accurate object segmentation, and then associate each object segment with either one of the detected object labels or none.

To segment the depth image into objects, we follow the unsupervised method proposed in \cite{Pham16b} with improvements. We first over-segment the depth image into supervoxels, and construct an adjacency graph connecting nearby supervoxels. The object segmentation task is then formulated as partitioning the graph into connected components. However, unlike \cite{Pham16b} where the graph edges are cut heuristically by classifying edges into either 1 (kept) or 0 (cut), we place a weight on each edge, and find the optimal cut using Kruskal's algorithm \cite{Felzenszwalb2004}. Edges between supervoxels in the same object should have low weights, and high weights otherwise. Let $e_{ij}$ be an edge connecting two supervoxels $s_i$ and $s_j$ with normals $n_i$ and $n_j$ respectively, its weight is calculated as:
\begin{align}
  w(e_{ij}) =  \begin{cases} 0~\mbox{if $s_i$ and $s_j$ on the same supporting plane,}\\
                             1~\mbox{if either $s_i$ or $s_j$ on a supporting plane,}\\
                             (1- n_i \cdot n_j)^2 ~\mbox{if $s_i$ and $s_j$ are convex,}\\
                             (1- n_i \cdot n_j) ~~\mbox{if $s_i$ and $s_j$ are concave.}
                            \end{cases}\nonumber
\end{align}
We refer readers to \cite{Pham16b} for technical details of extracting supporting planes in 3D scenes, and \cite{Stein2014} for convex/concave calculations. When executed, the 3D segmentation typically takes 175 ms.

\subsection{Data Association}

After the segmentation assigned a 3D point cloud to every object detection, the data association step determines whether the detected object is already in the map or needs to be added as a new map object. We perform this by a two-stage pipeline: first for every detection, a set of candidate object landmarks are selected based on the Euclidean distance of the respective point cloud centroids. 
Then we perform a nearest neighbor search between the 3D points in the landmark and in the detection, and calculate the Euclidean distance between the associated point pairs. A k-d tree structure helps to keep this step efficient. A detection is associated to an existing landmark if at least $50\%$ of its 3D points have a distance of 2 cm or less. This data association step typically takes around 30 ms per detection-landmark pair.

\subsection{Object Model Update}

As illustrated in Fig. \ref{fig:system_overview}, every object in our map maintains (i) the segmented colored 3D point clouds associated with that object by the data association step, (ii) an index vector into the pose variables of ORB-SLAM's factor graph, corresponding to the poses the landmark was observed from, and (iii) the accumulated per-class confidences provided by the SSD object detector. The latter is a vector $\vs$ of length $\|\cC\|$, where $\cC$ is the set of known classes.
Whenever a detection is associated with a map object, $\vs$ is updated according to $s_c = s_c + \mathfrak{s}$, 
where $c$ and $\mathfrak{s}$ are the class ID reported by SSD, and the associated confidence.
The class label for an object is determined by the accumulated score $\argmax_c s_c$ and a final confidence $\sigma$ can be assigned as $\sigma = \max_c s_c / n$ where $n$ is the total number of observations for that object.

Storing the point cloud segments with every object allows us to re-build an object model when the SLAM system updates its trajectory estimate, e.g. after a loop closure. For data efficiency, we store downsampled point clouds with 5 mm spatial resolution. This sparse model is also used during the data association step described above.

\subsection{Map Generation}
The map in our system is maintained implicitly by storing (i) the 3D point clouds observed by the camera at every keyframe, and (ii) storing the segmented 3D point clouds for every object, along with a pointer into ORB-SLAM's pose graph. If the map (or local subsets of it) is needed explicitly, for instance for path planning or grasp point selection, we can generate it by projecting the stored 3D points according to the current best estimate of the associated poses. When creating the map this way, we maintain a resolution of 1 cm for non-objects and 0.5 cm for objects. Depending on the application requirements, this sparse point cloud map can be turned into a dense mesh by appropriate algorithms. 

To make sure the data association has the most up-to-date information, we update the point cloud models of an object whenever it is observed.

\begin{figure*}
    \centering
    \includegraphics[width=0.32\linewidth]{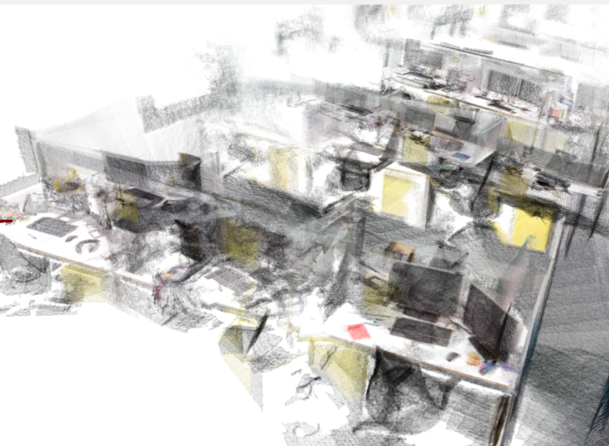}
    \includegraphics[width=0.32\linewidth]{S1130_phd-seg.png}
    \includegraphics[width=0.32\linewidth]{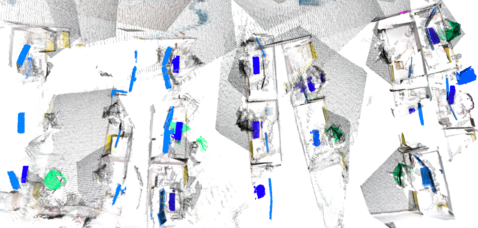}
    \caption{Semantic mapping in a lab-sized environment: Here we walked through a large combined office and lab space, without sweeping the camera closely to the desks. There are 36 individual monitors in this environment, but the system reported only 31 (light blue), since it mapped some dual-monitor setups as a single monitor.}
    \label{fig:S1130-phd}
\end{figure*}

\begin{figure*}
    \centering
    \includegraphics[width=0.24\linewidth]{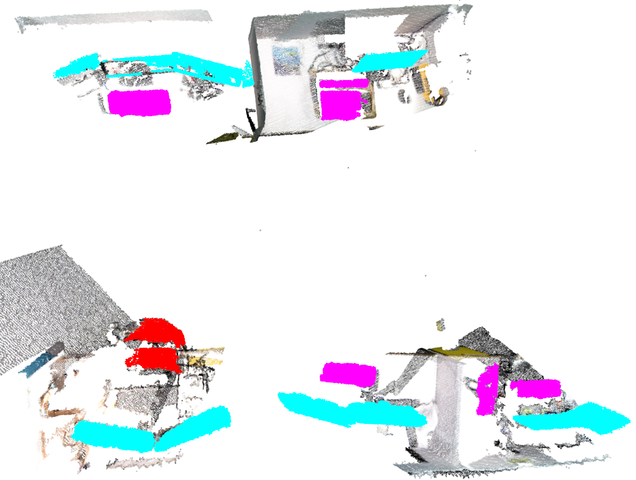}
    \includegraphics[width=0.24\linewidth]{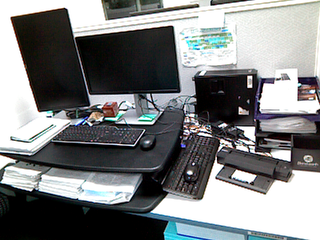}
    \includegraphics[width=0.24\linewidth]{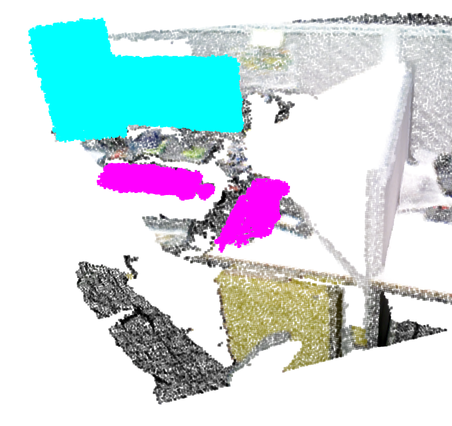}
    \includegraphics[width=0.24\linewidth]{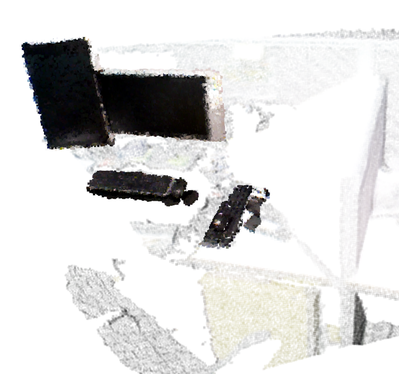}
    \caption{Mapping cluttered office scenes: The created map (left picture) shows our system correctly identified and mapped all 10 monitors (cyan), and detected 5 of the 6 keyboards (pink). The 2nd image from the left shows a close-up of the desk in the lower right of the map. Despite the clutter, the two monitors and keyboards were correctly detected and mapped (2nd from right). The rightmost image shows the geometric object models projected into the map model.}
    \label{fig:desk1}
\end{figure*}

\begin{figure*}
    \centering
    \includegraphics[width=0.32\linewidth]{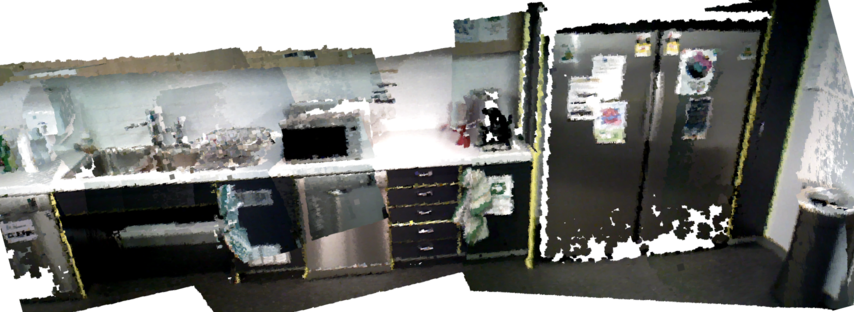}
    \includegraphics[width=0.32\linewidth]{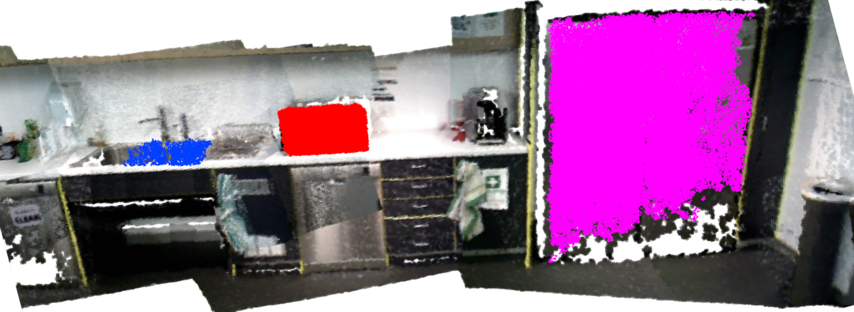}
    \includegraphics[width=0.32\linewidth]{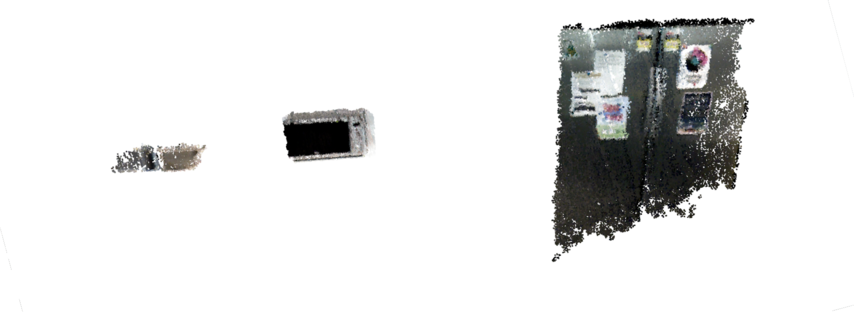}
    \caption{Two individual sinks (blue), the microwave(red), and the fridge (pink) have been mapped in this kitchen sequece. Full reconstructed 3D point cloud map (left), superimposed class labels (middle), and object models only, projected into their estimated pose in the world (right).}
    \label{fig:cantina}
\end{figure*}

\begin{figure*}
    \centering
    \includegraphics[width=0.32\linewidth]{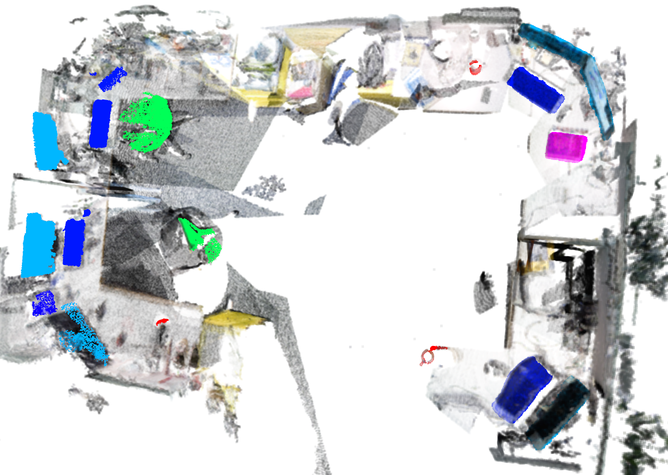}
    \includegraphics[width=0.32\linewidth]{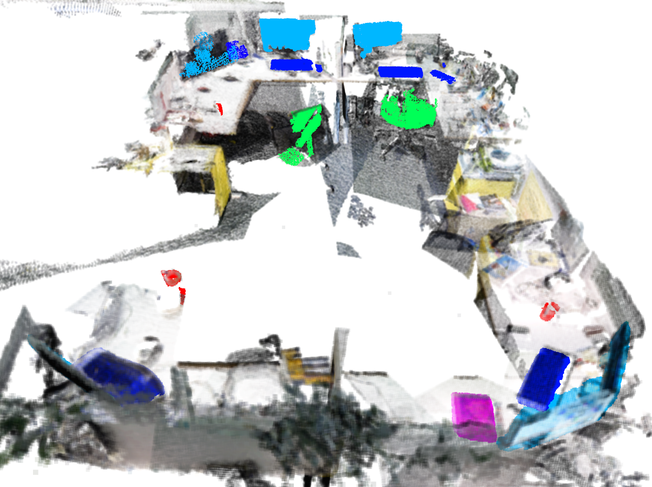}
     \includegraphics[width=0.32\linewidth]{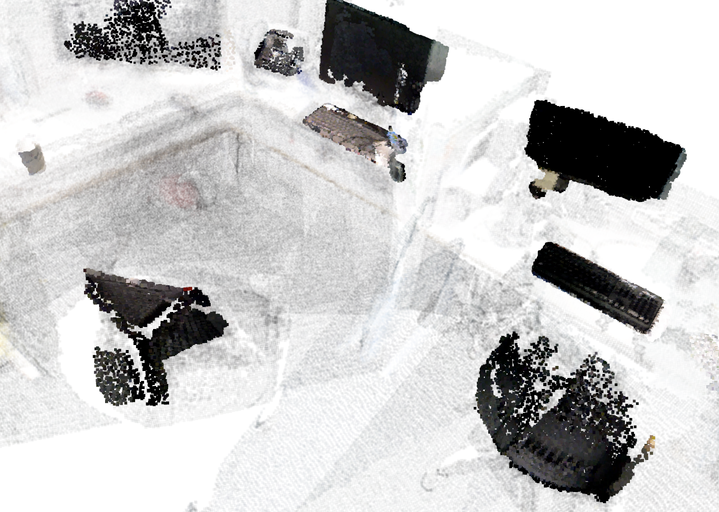}
    \caption{This cluttered office sequence contains numerous monitors (light blue), keyboards (dar blue), books (pink), cups (red), and chairs (green). The rightmost panel shows a close up of the left part of the map, with the geometric point cloud models of the mapped objects projected into the general non-object map.}
    \label{fig:S1130-office}
\end{figure*}

\section{Evaluation and Lessons Learned}
\label{sec:results}
We demonstrated and evaluated the capabilities of our object-oriented semantic mapping system by creating semantic maps in indoor environments of different scale, ranging from a single desk, a larger office, a kitchen, to a complete lab space. We generated a full map of each environment and compared the number of mapped object instances per class with the actual (ground truth) quantity of objects visible in the sequence. The results are summarized in Table~\ref{tab:inventory}, while Figures \ref{fig:desk1} - \ref{fig:cantina} show the resulting maps and highlight a number of details. 

Our system is able to correctly identify the majority of objects of interest in the evaluated scenes. As shown in Table~\ref{tab:inventory}, we observed only two false positive detections (i.e. mapping an object although it was not there in reality). One occurred in the office sequence when the lower corner of a window was mistaken for a monitor by the SSD object detector, the other occurred when a monitor appeared twice in the map due to errors in the depth perception. 

The more prominent failures are false negative detections, i.e. the system fails to map existing objects. False negatives are either caused because objects are not detected by the object detection, or the detections are discarded by later processing steps in the pipeline illustrated in Fig. \ref{fig:system_overview}. We address each of the observed causes for failures separately and point out directions for future research that can help to overcome the encountered challenges.

\paragraph{Failing Instance Segmentation}
In the Lab sequence, 5 monitors failed to be mapped. This is due to the dual-monitor setup used on most desks; some of those monitor pairs could not be separated into two distinct objects by the segmentation approach, due to their near parallel alignment. Failure cases such as this can be mitigated by extending the semantic mapping system with capabilities to reason over the spatial structure and sizes of objects. In that particular example, prior knowledge about the typical dimensions of monitors encountered in an office environment would support the hypothesis of actually observing two separate monitors, not just one. Such prior knowledge could also be obtained from annotated training data.

\paragraph{Corrupted Depth Perception}
In the Office sequence, a highly reflecting iMac disrupted the depth perception and subsequently led to failures in the segmentation and data association. This failure then resulted in two independent objects being mapped, instead of only one. Recent advances in the field of single-view depth estimation with CNNs (e.g., \cite{Eigen2014, Garg16}) that implicitly exploit semantic knowledge to determine the most likely depth structure of a scene can be adopted to correct such failures. 

Of the 30 keyboards present in all sequences, 8 were missed by our mapping system. These failures were caused by noisy depth perception that complicated the reliable segmentation of the flat keyboards on the surface of the desk. An instance-level 3D segmentation approach that better exploits visual appearance or mid-level convolutional features would mitigate such effects.

\paragraph{Training Set Discrepancies}
The books in our test data were particularly hard for the Convolutional Network used by SSD to detect reliably. A discrepancy in spatial orientation with respect to the camera, and appearance variations between SSD's training dataset (MS COCO~\cite{Lin14}) and our real-world test scenes explains these mistakes. This failure illustrates one of the major issues currently facing robotics: how well do typical computer vision datasets such as ImageNet~\cite{Russakovsky15} or COCO accurately represent the environmental and appearance conditions encountered in robotic "in the wild" scenarios?

Although SSD was trained on the COCO dataset that contains 80 distinct classes, only a small subset of 10 of them actually appeared in the tested indoor environments (backpack, chair, keyboard, laptop, monitor, computer mouse, cell phone, sink, refrigerator, microwave). The majority of COCO classes comprise animals and objects typically encountered outdoors. For applications of robotic semantic mapping, it is likely that further improvements can be achieved by applying incremental and low-shot learning techniques (e.g., \cite{Bertinetto2016}) to extend the recognition capabilities of the classifier, instead of using a generic pre-trained network.

\paragraph{Low-Resolution Cameras}
The obtained results also indicate that mapping of small objects is particularly challenging. This difficulty stems partly from the low resolution (both in RGB and in the depth) of the used PrimeSense RGB-D sensor. Small objects typically have a low chance of being detected by SSD, especially on the noisy and often blurry images of the camera. Even if such objects are detected in the image, they contain only a few 3D points  and thus segmentation, data association and alignment are not reliable and these detections are most often discarded. For practical implementations, the obvious solution is to use higher resolution sensors and the technology is still developing rapidly, meaning this should become a practical in the near future at the same price.

\begin{table}
\centering
\begin{tabular}{@{}lllll@{}}
    \toprule
     Sequence & Objects &  \multicolumn{3}{c}{Mapped}   \\ 
     & & true pos & false pos & false neg  \\ \midrule
     Lab (Fig. \ref{fig:S1130-phd})& monitors & 31 & 0 & 5  \\          
         & keyboards & 12 & 0 & 7  \\
     Desk (Fig. \ref{fig:desk1}) & keyboards & 5 & 0 & 1 \\
        & monitors & 10 & 0 & 0  \\
     Kitchen (Fig. \ref{fig:cantina}) & sinks & 2 & 0 & 0 \\
             & microwave & 1 & 0 & 0 \\      
             & fridge & 1  & 0 & 0 \\      
     Office (Fig. \ref{fig:S1130-office})& monitors & 6  & 2 & 0 \\      
            & keyboards & 5  & 0 & 0\\
            & cups & 3  & 0 & 0\\
            & chair & 3  & 0 & 1\\
            & books & 1 & 0 & $>10$ \\
            & telephone & 1 & 0 & 2 \\
     \bottomrule
\end{tabular}
\caption{Quantitative results of our semantic mapping approach, listing true positive detections (true pos), missed objects (false neg), and mis-detections (false pos). See the text for a discussion of these results, and the failure cases in particular.}
\label{tab:inventory}
\end{table}

\section{Conclusions and Future Work}

We presented a novel combination of SLAM, object detection, instance-level segmentation, data association, and model updates to obtain a semantic mapping system that maintains individual objects as the key entities in the map. Our approach differs from previous approaches in that it builds 3D object models on the fly, does not require \textit{a-priori} known 3D models, and can leverage the full potential of deep-learnt object detection methods. We demonstrated and evaluated the efficacy of this approach in an automated inventory management scenario by mapping and semantically annotating numerous indoor scenes in a typical workplace office environment.

We discussed the observed failure cases and proposed directions for future work to address them. In addition, we will investigate how the detected objects can serve as semantic landmarks to improve the accuracy of the SLAM system, thus closing the loop to create a full semantic SLAM system. This avenue of investigation also leads to the question of how an image-based object detector like SSD and other deep-learnt approaches can be best treated as a sensor and tightly integrated into the data fusion framework of factor graphs that are commonly applied as backends in SLAM.
Furthermore, the objects in our system are currently represented as collections of point clouds. In future work we are going to utilize methods like \cite{Whelan15} to obtain dense surface models.
The proposed future research can be supported by recently published synthetic datasets \cite{Mccormac16} or high-fidelity simulation environments \cite{Qiu16}.

Investigating how semantic maps can benefit other task domains like robotic planning for mobile manipulation, path planning or general behaviour generation will yield more insights into what level (or levels) of semantic representations are appropriate in different application domains.

\bibliographystyle{IEEEtran}
\bibliography{bibfile}

\end{document}